\documentclass[conference]{IEEEtran}
\IEEEoverridecommandlockouts

\usepackage{cite}
\usepackage{amsmath,amssymb,amsfonts}
\usepackage{algorithmic}
\usepackage{graphicx}
\usepackage{textcomp}
\usepackage{mydef}
\usepackage{hyperref}
\usepackage{xcolor}
\usepackage{tikz}
\usepackage[edges]{forest}
\definecolor{tablebordercolor}{rgb}{0.373, 0.620, 0.627} 
\def\BibTeX{{\rm B\kern-.05em{\sc i\kern-.025em b}\kern-.08em
    T\kern-.1667em\lower.7ex\hbox{E}\kern-.125emX}}
\begin{document}
\newcommand{\binzhu}[1]{{\color{pink}[binzhu: #1]}}

\title{Advancing Extended Reality with 3D Gaussian Splatting: Innovations and Prospects
\thanks{This work was supported in part by The Chinese University of Hong Kong (Project No.: 4055212); and in part by the Research Grants Council of the Hong Kong Special Administrative Region, China (Project No.: T45-401/22-N).}
\thanks{\{shiqiu, bzxie, qxliu, pheng\}@cse.cuhk.edu.hk}
}

\author{\IEEEauthorblockN{{Shi Qiu, Binzhu Xie, Qixuan Liu, Pheng-Ann Heng}}\\
\IEEEauthorblockA{\textit{Department of Computer Science and Engineering, The Chinese University of Hong Kong, HKSAR China} \\
\textit{Institute of Medical Intelligence and XR, The Chinese University of Hong Kong, HKSAR China}
}
}

\maketitle

\begin{abstract}
3D Gaussian Splatting (3DGS) has attracted significant attention for its potential to revolutionize 3D representation, rendering, and interaction. Despite the rapid growth of 3DGS research, its direct application to Extended Reality (XR) remains underexplored. Although many studies recognize the potential of 3DGS for XR, few have explicitly focused on or demonstrated its effectiveness within XR environments. In this paper, we aim to synthesize innovations in 3DGS that show specific potential for advancing XR research and development. We conduct a comprehensive review of publicly available 3DGS papers, with a focus on those referencing XR-related concepts. Additionally, we perform an in-depth analysis of innovations explicitly relevant to XR and propose a taxonomy to highlight their significance. Building on these insights, we propose several prospective XR research areas where 3DGS can make promising contributions, yet remain rarely touched. By investigating the intersection of 3DGS and XR, this paper provides a roadmap to push the boundaries of XR using cutting-edge 3DGS techniques.
\end{abstract}

\begin{IEEEkeywords}
3D Gaussian Splatting, Neural Rendering, XR.
\end{IEEEkeywords}

\section{Introduction}
Extended Reality (XR), which includes Virtual Reality (VR), Augmented Reality (AR), and Mixed Reality (MR), represents a frontier in immersive technologies, aiming to seamlessly blend the physical and digital worlds. Achieving truly immersive XR experiences relies on advancements in 3D representation, rendering, and interaction techniques. In recent years, Neural Radiance Fields (NeRF)~\cite{mildenhall2021nerf} have led the way in neural rendering research, enabling photorealistic depiction of complex scenes from multi-view observations. However, NeRF-based methods~\cite{pumarola2021d,barron2021mip,wang2021neus} often face challenges in real-time performance and effective integration into XR systems due to their high computational complexity and reliance on implicit scene representations through Multi-Layer Perceptrons (MLPs). This implicit representation, while powerful, makes it difficult to directly manipulate or interact with the underlying 3D geometry, further complicating their integration into XR systems. Recently, 3D Gaussian Splatting (3DGS)~\cite{kerbl20233d} has emerged as a promising alternative for effective 3D representations. By explicitly modeling scenes using learnable, parameterized 3D Gaussians, real-time rendering and manipulation can be achieved, offering a significant improvement in performance over NeRF approaches. The inherent efficiency and flexibility of 3DGS make it an ideal candidate for XR, where both responsive interaction and synergistic system integration are essential.

\begin{figure}
\begin{center}
\includegraphics[width=0.9\columnwidth]{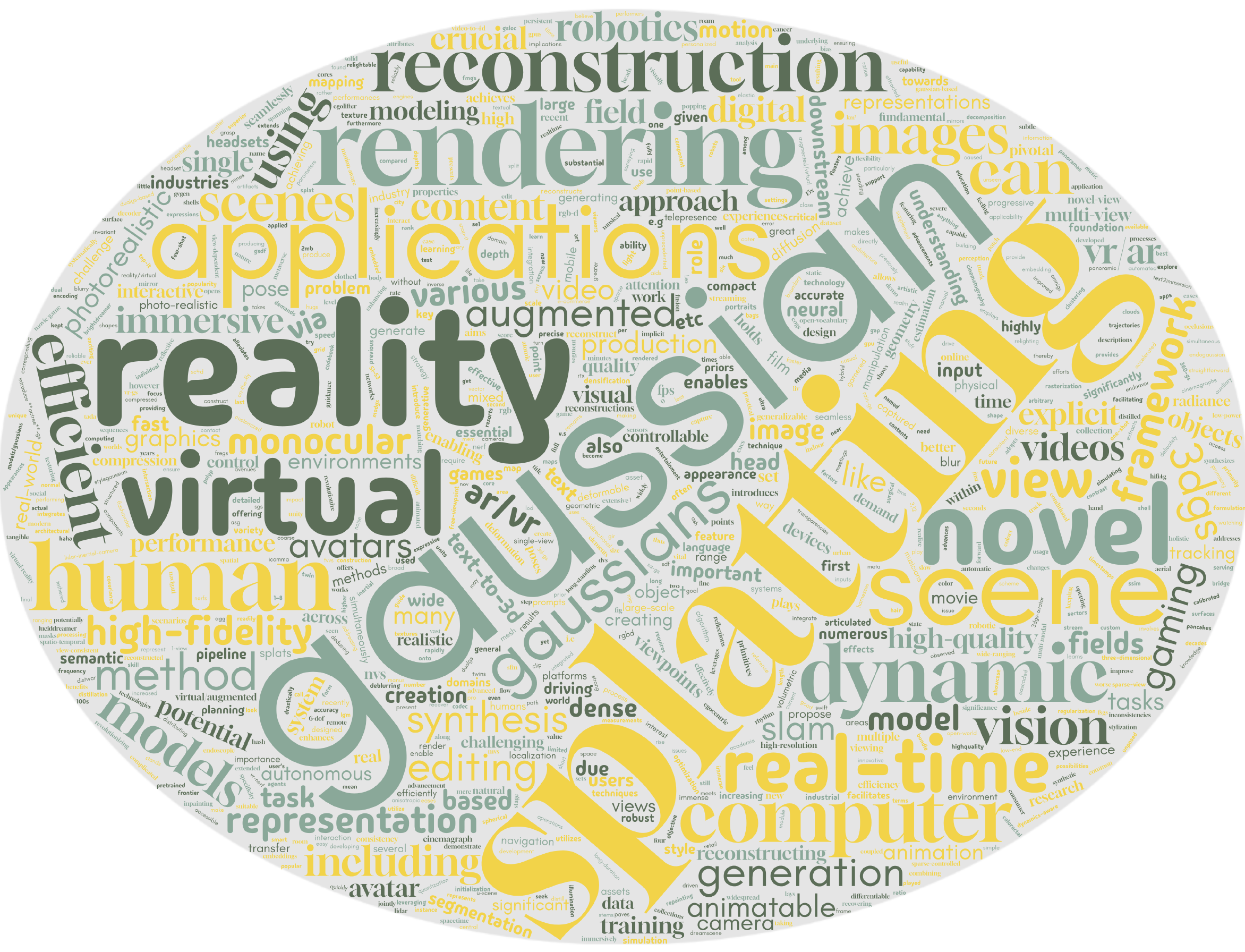}
\end{center}
   \caption{Word cloud extracted from recent 3DGS literature that references XR-related keywords. The size of each word reflects its frequency of occurrence, highlighting key concepts and trends in the intersection of 3DGS and XR technologies.}
\label{fig:word}
\end{figure}

Despite rapid advancements and an increasing number of studies on 3DGS, its direct application within XR environments remains notably underexplored. To address this gap, we conduct a comprehensive review of the 3DGS literature, based on 272 publicly available papers\footnote{Our collected papers span from the very first 3DGS publication~\cite{kerbl20233d} in July 2023 to those available as of late-October 2024.} that cover recent published works in top-tier AI venues such as CVPR, ECCV, SIGGRAPH, and NeurIPS, \emph{etc}. We systematically examine each paper, screening for XR-related content to assess their relevance to XR domains. Our preliminary analysis shows that more than half of these papers (\emph{i.e.}, 152 out of 272) explicitly mention XR concepts. Fig.~\ref{fig:word} presents a word cloud visualizing the frequency of XR-related terms within the collected papers, highlighting common associations such as ``virtual reality'', ``novel scene'', ``rendering'', ``reconstruction'', \emph{etc}.

Although many studies acknowledge the potential relevance of 3DGS to XR, our investigation reveals that the majority \emph{do not} specifically exploit 3DGS techniques in XR environments. Instead, they often reference XR in a general and superficial manner, mentioning it as a potential application area without providing specific XR-based implementations or evaluations. The lack of studies that concretely apply 3DGS in XR settings presents both a challenge and an opportunity: the challenge of effectively integrating emerging 3DGS innovations into XR systems, and the opportunity to enhance XR experiences through the state-of-the-art 3DGS techniques. In this paper, we synthesize existing innovations, explore opportunities for future research and development, and provide a roadmap for exploiting the synergies between 3DGS and XR technologies. By doing so, we aim to inspire and guide future efforts in leveraging 3DGS to push the boundaries of XR, contributing to more immersive, interactive, and intelligent experiences. 

\tikzstyle{leaf}=[draw=hiddendraw,
    rounded corners,minimum height=1em,
    fill=hidden-orange!40,text opacity=1, align=center,
    fill opacity=.5,  text=black,align=center,font=\scriptsize,
    inner xsep=3pt,
    inner ysep=2.5pt,
    ]
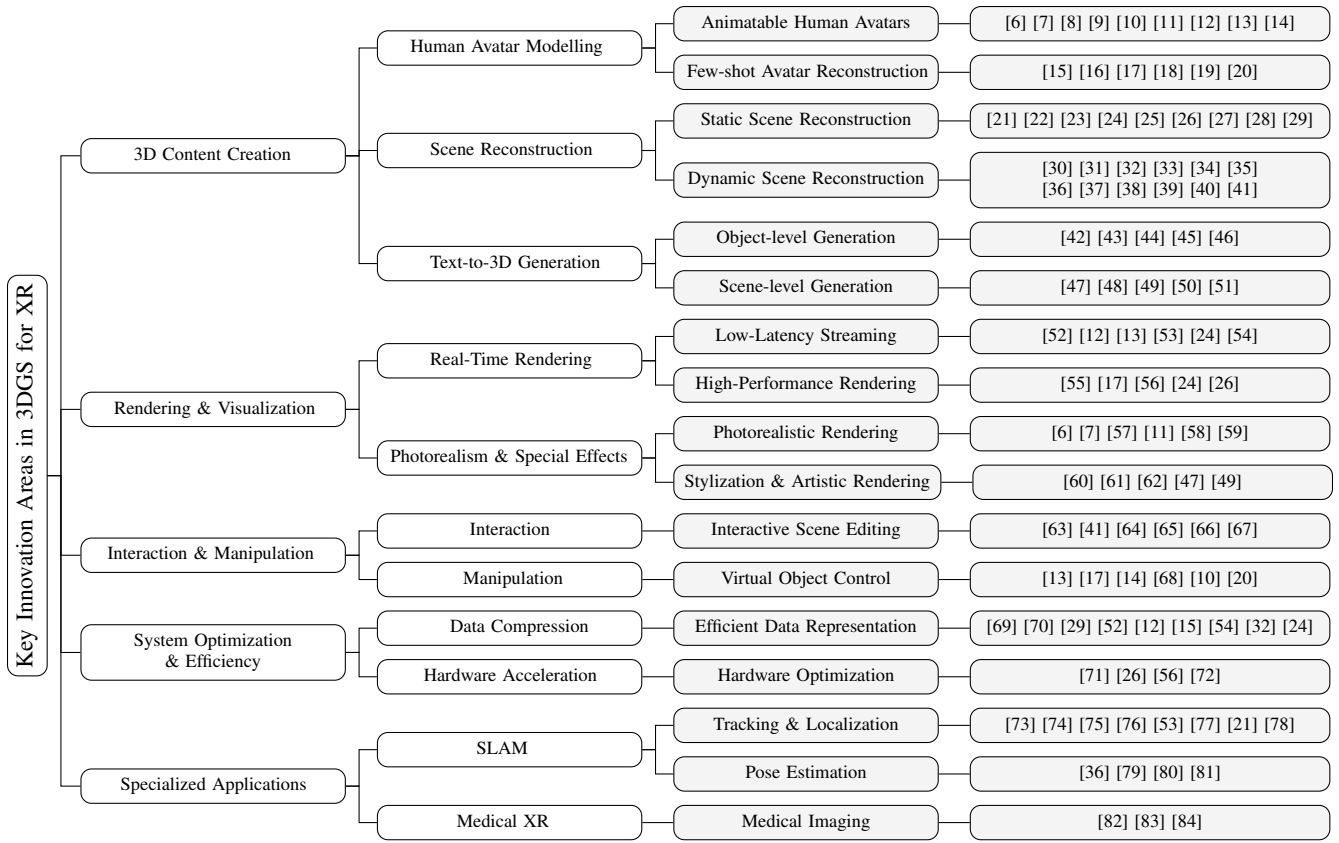
\begin{figure*}[ht]
\centering
\begin{forest}
  for tree={
  forked edges,
  grow=east,
  reversed=true,
  anchor=base west,
  parent anchor=east,
  child anchor=west,
  base=middle,
  font=\scriptsize,
  rectangle,
  draw=hiddendraw,
  rounded corners,align=center,
  minimum width=10em,
    s sep=5pt,
    inner xsep=3pt,
    inner ysep=2.5pt,
  },
  where level=1{text width=4.5em}{},
  where level=2{text width=6em,font=\scriptsize}{},
  where level=3{font=\scriptsize}{},
  where level=4{font=\scriptsize}{},
  where level=5{font=\scriptsize}{},
  [Key Innovation Areas in 3DGS for XR,rotate=90,anchor=north,edge=hiddendraw, font=\small
    [3D Content Creation,edge=hiddendraw,align=center,text width=6em
        [Human Avatar Modelling, text width=7.5em, edge=hiddendraw
            [Animatable Human Avatars,leaf,text width=8em, edge=hiddendraw
                        [\cite{moreau2024human}\cite{pang2024ash}\cite{yuan2024gavatar}\cite{hu2024gaussianavatar}\cite{dhamo2023headgas}\cite{yang2023real}\cite{jiang2024robust}\cite{jiang2024vr}\cite{saito2024relightable},leaf,text width=13em, edge=hiddendraw]
                        ]
            [Few-shot Avatar Reconstruction,leaf,text width=9em, edge=hiddendraw
                        [\cite{wen2024gomavatar}\cite{chu2024generalizable}\cite{li2023human101}\cite{qian20243dgs}\cite{xiang2024flashavatar}\cite{chen2024monogaussianavatar},leaf,text width=13em, edge=hiddendraw]
                        ]
        ]
        [Scene Reconstruction, text width=6em, edge=hiddendraw
            [Static Scene Reconstruction,leaf,text width=8em, edge=hiddendraw
                                    [\cite{hong2024liv}\cite{lin2024vastgaussian}\cite{bai2024360}\cite{radl2024stopthepop}\cite{lee2024deblurring}\cite{lee2024gscore}\cite{yu2024mip}\cite{lu2024scaffold}\cite{liu2024citygaussian},leaf,text width=13em, edge=hiddendraw]
            ]
            [Dynamic Scene Reconstruction,leaf,text width=9em, edge=hiddendraw
                                    [\cite{duan20244d}\cite{yang2024deformable}\cite{katsumata2023efficient}\cite{liu2024modgs}\cite{lu20243d}\cite{zhu2024motiongs}\\\cite{luiten2023dynamic}\cite{bae2024per}\cite{lin2024gaussian}\cite{li2024spacetime}\cite{kratimenos2023dynmf}\cite{huang2024sc},leaf,text width=13em, edge=hiddendraw]
                                    ]
        ]
        [Text-to-3D Generation, text width=6em, edge=hiddendraw
            [Object-level Generation,leaf,text width=8em, edge=hiddendraw
                                    [\cite{tang2025lgm}\cite{he2024gvgen}\cite{liang2024luciddreamer}\cite{wang20243d}\cite{qin2024langsplat},leaf,text width=13em, edge=hiddendraw]
            ]
            [Scene-level Generation,leaf,text width=9em, edge=hiddendraw
                                    [\cite{li2024dreamscene}\cite{zhou2025dreamscene360}\cite{ouyang2023text2immersion}\cite{shriram2024realmdreamer}\cite{zhou2024gala3d},leaf,text width=13em, edge=hiddendraw]
                                    ]
        ]
    ]
    [Rendering \& Visualization,edge=hiddendraw,align=center,text width=7.5em
        [Real-Time Rendering, text width=6em, edge=hiddendraw
            [Low-Latency Streaming,leaf,text width=8.5em, edge=hiddendraw
                        [\cite{jiang2024hifi4g}\cite{jiang2024robust}\cite{jiang2024vr}\cite{peng2024rtg}\cite{radl2024stopthepop}\cite{sun20243dgstream},leaf,text width=13em, edge=hiddendraw]
                        ]
            [High-Performance Rendering,leaf,text width=9em, edge=hiddendraw
                        [\cite{zhou2024headstudio}\cite{li2023human101}\cite{niedermayr2024compressed}\cite{radl2024stopthepop}\cite{lee2024gscore},leaf,text width=13em, edge=hiddendraw]
                        ]
        ]
        [Photorealism \& Special Effects, text width=9em, edge=hiddendraw
            [Photorealistic Rendering,leaf,text width=8em, edge=hiddendraw
                                    [\cite{moreau2024human}\cite{pang2024ash}\cite{qian2024gaussianavatars}\cite{yang2023real}\cite{yang2024spec}\cite{liang2024gs},leaf,text width=13em, edge=hiddendraw]
            ]
            [Stylization \& Artistic Rendering,leaf,text width=9.5em, edge=hiddendraw
                                    [\cite{liu2024stylegaussian}\cite{saroha2024gaussian}\cite{li2024loopgaussian}\cite{li2024dreamscene}\cite{ouyang2023text2immersion},leaf,text width=13em, edge=hiddendraw]
                                    ]
        ]
    ]
    [Interaction \& Manipulation,edge=hiddendraw,align=center,text width=8em
        [Interaction, text width=3em, edge=hiddendraw
            [Interactive Scene Editing,leaf,text width=8em, edge=hiddendraw
                        [\cite{yu2024cogs}\cite{huang2024sc}\cite{wang2025view}\cite{jaganathan2024ice}\cite{chen2024gaussianeditor}\cite{wang2024gaussianeditor},leaf,text width=13em, edge=hiddendraw]
                        ]
        ]
        [Manipulation, text width=3.5em, edge=hiddendraw
            [Virtual Object Control,leaf,text width=8em, edge=hiddendraw
                                    [\cite{jiang2024vr}\cite{li2023human101}\cite{saito2024relightable}\cite{rivero2024rig3dgs}\cite{dhamo2023headgas}\cite{chen2024monogaussianavatar},leaf,text width=13em, edge=hiddendraw]
            ]
        ]
    ]
    [System Optimization \\\& Efficiency,edge=hiddendraw,align=center,text width=6em
        [Data Compression, text width=4.5em, edge=hiddendraw
            [Efficient Data Representation,leaf,text width=8.5em, edge=hiddendraw
                        [\cite{fan2023lightgaussian}\cite{abdal2024gaussian}\cite{liu2024citygaussian}\cite{jiang2024hifi4g}\cite{jiang2024robust}\cite{wen2024gomavatar}\cite{sun20243dgstream}\cite{katsumata2023efficient}\cite{radl2024stopthepop},leaf,text width=13em, edge=hiddendraw]
                        ]
        ]
        [Hardware Acceleration, text width=6.5em, edge=hiddendraw
            [Hardware Optimization,leaf,text width=8em, edge=hiddendraw
                                    [\cite{navaneet2024compgs}\cite{lee2024gscore}\cite{niedermayr2024compressed}\cite{yu2024gsdf},leaf,text width=13em, edge=hiddendraw]
            ]
        ]
    ]
    [Specialized Applications,edge=hiddendraw,align=center,text width=7em
        [SLAM, text width=2.5em, edge=hiddendraw
            [Tracking \& Localization,leaf,text width=8em, edge=hiddendraw
                                    [\cite{ji2024neds}\cite{sun2024high}\cite{hu2024cg}\cite{sun2024mm3dgs}\cite{peng2024rtg}\cite{yan2024gs}\cite{hong2024liv}\cite{keetha2024splatam},leaf,text width=13em, edge=hiddendraw]
            ]
            [Pose Estimation,leaf,text width=9em, edge=hiddendraw
                                    [\cite{luiten2023dynamic}\cite{pokhariya2024manus}\cite{cai2024gs}\cite{li2024ggrt},leaf,text width=13em, edge=hiddendraw]
                                    ]
        ]
        [Medical XR, text width=4em, edge=hiddendraw
            [Medical Imaging,leaf,text width=9em, edge=hiddendraw
                        [\cite{liu2024endogaussian}\cite{bonilla2024gaussian}\cite{niedermayr2024application},leaf,text width=13em, edge=hiddendraw]
                        ]
        ]
    ]
    ]
\end{forest}
\caption{Taxonomy of 3DGS Innovations Relevant to XR.}
\label{taxonomy}
\end{figure*}

\section{Related Works}
\subsection{Representative 3DGS Works Targeting XR}
Unlike most existing 3DGS studies that superficially mention XR, we have identified three representative works that specifically leverage 3DGS to advance XR research and development.

\textbf{VR-GS}~\cite{jiang2024vr}: Interacting with 3D models in VR is time-consuming and complex due to the need for real-time synchronization of distinct geometry representations for simulation and rendering. Jiang~\emph{et al.} introduce \emph{VR-GS}, a physics-aware system based on 3DGS for smooth creation, editing, and interaction experiences in VR~\cite{jiang2024vr}. (i) To ensure real-time deformation, VR-GS employs eXtended Position-based Dynamics (XPBD) \cite{macklin2016xpbd}. (ii) To address the geometry mismatch between simulation and rendering, VR-GS constructs tetrahedral cages embedding GS kernel groups, driven by XPBD to guide deformation. By using a two-level embedding to prevent undesirable deformations and enabling offline preparation, VR-GS is ideal for VR content editing, significantly enhancing the immersive experience with realistic dynamics.

\textbf{DualGS}~\cite{jiang2024robust}: Volumetric videos provide a 6-DoF immersive experience in XR, but their creation, storage, and rendering often demand extensive manual intervention. Jiang~\emph{et al.} reduce storage and rendering demands for volumetric videos by introducing \emph{Dual {G}aussian {S}platting} (DualGS) to realize efficient, high-fidelity real-time rendering of human performance in XR environments~\cite{jiang2024robust}. (i) To reduce motion redundancy and enhance temporal coherence, DualGS decouples motion and appearance using joint and skin Gaussians \cite{loper2023smpl}, employing a coarse-to-fine optimization to improve temporal coherence and tracking. (ii) For the compression strategy, DualGS uses entropy encoding and lookup table compression to reduce storage to $\sim$350KB per frame, realizing immersive user experiences like feeling present at a musician's concert in XR.

\textbf{RGCA}~\cite{saito2024relightable}: Virtual avatar rendering shows great potential for XR-based gaming and telecommunication. To tackle lighting and detail challenges in virtual avatars for enhanced realism, Saito~\emph{et al.} introduce \emph{Relightable Gaussian Codec Avatars} (RGCA), leveraging 3D Gaussians for high-fidelity, real-time rendering under various lighting conditions~\cite{saito2024relightable}. (i) To overcome the challenges of light scattering and reflection in complex materials: RGCA presents a relightable appearance model with learnable radiance transfer, utilizing spherical harmonics and Gaussians; and invents a novel parameterization for specular reflection to accurately approximate complex light effects. (ii) For accurate hair rendering, RGCA uses convolutional layers to encode driving signals that captures fine hair details. (iii) For eye representation, RGCA develops an explicit eye model to improve realism and gaze control. These innovations tackle the complexities and challenges of human head rendering in XR applications.

\subsection{3DGS Surveys}
In this year of 2024, several surveys~\cite{bai2024progressprospects3dgenerative,chen2024survey3dgaussiansplatting,Fei_2024,tosi2024nerfs3dgaussiansplatting,wu2024recentadvances3dgaussian,Dalal_2024} have comprehensively explored the latest advancements from various perspectives of the 3DGS technology. Specifically, \cite{chen2024survey3dgaussiansplatting,Fei_2024} conduct systematic analyses of 3DGS, covering its foundational theories, optimization methods, applications, and future research directions. In addition, \cite{wu2024recentadvances3dgaussian} reviews the rapid progress of 3DGS methods, classifying them into 3D reconstruction, 3D editing, and other downstream applications. Furthermore, some reviews focus on specific downstream applications of 3DGS: (i) \cite{bai2024progressprospects3dgenerative} provides a detailed review of 3DGS development in 3D human modeling; (ii) \cite{tosi2024nerfs3dgaussiansplatting} explores the evolution of simultaneous localization and mapping (SLAM) technology, highlighting its recent progress in radiance fields including 3DGS; and (iii) \cite{Dalal_2024} concentrates on the latest techniques in 3D reconstruction and novel view synthesis, where 3DGS also plays an important role. While these surveys recognize the potential connections between 3DGS and XR, they do not specifically elaborate on their synergies and prospects, which are the primary focus of this paper. 





\section{Innovations in 3D Gaussian Splatting for XR}
To understand the impact of 3DGS on XR related areas, we have analyzed the collected literature and key innovations relevant to XR. As illustrated in Fig.~\ref{taxonomy}, we identify five key areas where innovations significantly align with XR applications, forming the basis of our proposed taxonomy.

\textbf{{3D Content Creation:}}
In \emph{human avatar modeling}, 3DGS can enhance XR applications by enabling rapid, high-fidelity creation of animatable avatars, which are essential for immersive and personalized XR experiences in telepresence, virtual meetings, and interactive VR~\cite{yuan2024gavatar,hu2024gaussianavatar,yang2023real}. 3DGS also facilitates the reconstruction of realistic human avatars from minimal input data, such as monocular videos, directly addressing the demand driven by AR/VR wearables~\cite{wen2024gomavatar,chen2024monogaussianavatar,li2023human101}. Additionally, 3DGS can support real-time rendering of high-quality human performances on VR headsets, enhancing accessibility and user engagement in XR collaborations \cite{moreau2024human,pang2024ash,dhamo2023headgas}. In \emph{scene reconstruction}, 3DGS provides efficient modeling of large-scale and complex scenes, which is essential for immersive navigation and interaction in virtual environments~\cite{lin2024vastgaussian,liu2024citygaussian}. 3DGS also allows for real-time rendering of extensive scenes from sparse views, facilitating free-viewpoint exploration and virtual tours in VR~\cite{bai2024360,lu2024scaffold,li2024spacetime}. In \emph{text-to-3D generation}, 3DGS simplifies the pipeline of creating immersive virtual contents from textual descriptions~\cite{tang2025lgm,liang2024luciddreamer,zhou2025dreamscene360}. This capability enhances interactivity and personalization by letting users create and modify virtual contents using natural language input~\cite{qin2024langsplat}. By addressing challenges related to the cost and complexity of developing rich digital worlds, 3DGS contributes to dynamic, user-adaptive functions, as well as more accessible and customizable XR experiences.

\textbf{{Rendering \& Visualization:}}
In \emph{real-time rendering}, 3DGS optimizes neural rendering pipelines to facilitate real-time visualization of complex scenes and models, which is critical for interactive XR experiences where latency affects immersion~\cite{peng2024rtg,niedermayr2024compressed}. By efficiently managing streaming resources, 3DGS can enable smooth rendering and visualization of 3D objects on a range of devices, including those with limited hardware capabilities such as mobile AR glasses and VR headsets~\cite{jiang2024robust,sun20243dgstream,jiang2024hifi4g}. In \emph{photorealism \& special effects}, 3DGS can improve visual fidelity by accurately modeling material properties and light interactions for high levels of realism in immersive virtual environments~\cite{qian2024gaussianavatars,liang2024gs}. Additionally, it supports the integration of artistic styles and visual effects, assisting in creating stylized or customized visual experiences in XR~\cite{liu2024stylegaussian,saroha2024gaussian}. 

\textbf{{Interaction \& Manipulation:}}
In \emph{interactive scene editing}, it is crucial for XR applications to empower users with the ability to interactively modify 3D elements in real-time within the virtual space. By allowing intuitive deformation and editing of 3D models through natural hand gestures or physics-aware interactions, 3DGS facilitates responsive object manipulation and environment customization, which are crucial for XR use cases like collaborative design, prototyping, and personalized scene creation~\cite{jiang2024vr,wang2025view,chen2024gaussianeditor,wang2024gaussianeditor}. In \emph{virtual object control,} users can leverage 3DGS to generate controllable portraits with detailed expressions and motions, realizing personalized avatar customization and dynamic expression rendering for engaging and responsive XR interactions.~\cite{dhamo2023headgas,chen2024monogaussianavatar,rivero2024rig3dgs}.

\textbf{{System Optimization \& Efficiency:}}
In \emph{data compression}, 3DGS optimizes the storage and rendering of complex 3D scenes by employing compact and efficient data structures, which is crucial for XR applications operating on mobile devices. The network pruning-driven method proposed in~\cite{fan2023lightgaussian} achieves up to 15x compression of 3D Gaussians while maintaining rendering speeds over 200 FPS. Furthermore, efficient 3D Gaussian representations for monocular or multi-view dynamic scenes can reduce memory usage and processing demands required for reconstruction, making immersive XR experiences more feasible on hardware with limited capabilities~\cite{katsumata2023efficient}. In \emph{hardware acceleration}, \cite{lee2024gscore} introduces specialized hardware and algorithmic accelerations to improve real-time 3DGS rendering on mobile devices, addressing computational challenges in resource-constrained XR applications. Additionally, \cite{niedermayr2024compressed,navaneet2024compgs} propose advanced optimization algorithms that further reduce memory usage and computational load of 3DGS techniques, aiming to achieve high-quality rendering on low-power devices such as AR/VR headsets. 

\textbf{{Specialized Applications:}}
In \emph{SLAM}, 3DGS serves as an efficient and effective 3D representation for real-time environment mapping and precise camera tracking. By leveraging 3DGS, dense visual SLAM can achieve improved rendering performance and trajectory accuracy, both essential for immersive XR experiences that depend on accurate spatial understanding~\cite{yan2024gs,sun2024high,sun2024mm3dgs}. Moreover, incorporating semantic and uncertainty-aware processing into 3DGS enhances robustness in dynamic environments, which can be further used in advanced XR applications like augmented reality navigation and interactive scanning~\cite{ji2024neds,hu2024cg}. In \emph{medical XR}, 3DGS leads to high-quality and real-time 3D reconstructions and visualizations of anatomical structures, which are crucial for medical diagnostics, surgical planning, and immersive training simulations. For instance, recent~\cite{liu2024endogaussian,bonilla2024gaussian} apply 3DGS to reconstruct detailed 3D surgical scenes from endoscopic videos, making them useful for surgical simulations and preoperative diagnosis. Furthermore, compressing large medical datasets into compact 3DGS representations can create interactive and photorealistic rendering of anatomical models on light XR devices, giving users access to examine detailed 3D anatomy in real time~\cite{niedermayr2024application}.

\section{Future Prospects}
Despite significant advancements in 3DGS research, most of the studies concentrate on regular neural rendering tasks, such as 3D reconstruction and generation. These developments have improved our ability to create high-quality visual content, building groundwork for XR applications. However, there still remain many opportunities to deeply explore the synergies between 3DGS and XR technologies. The future topics may include, but are not limited to, the following:

\begin{itemize}
    \item \emph{Mesh Modeling.} One fundamental 3DGS application in XR is mesh modeling. High-fidelity mesh modeling is essential for creating virtual environments where accurate interactions between users and digital objects are crucial~\cite{cui2022energy,zhu2024ssp}. By representing intricate surface details and complex geometries, we can further devise 3D Gaussians-based models to create more realistic and interactive 3D objects in virtual spaces. For example, GS-VTON~\cite{cao2024gs} enables VR content personalization based on user preferences.
    \item \emph{Dynamic Scene Representation.} 4D volumetric video is a valuable representation for dynamic scenes. To more effectively represent XR scenes, our future works should focus on enhancing 3DGS for real-time dynamic content capture, processing, and rendering with minimal latency. Moreover, efforts should target optimizing 3DGS to improve scalability for large-scale virtual spaces. For example, Gracia AI\footnote{\url{https://www.gracia.ai/}} develops a platform to create 3DGS-based volumetric videos for mobile VR devices.
    \item \emph{Open-world Scene Understanding.} In AR and MR domains, open-world scene understanding is critical for seamless integration of virtual content with the real world. 3DGS can make contributions to topic by leveraging open-vocabulary 3D segmentation, which allows systems to recognize and interpret various objects and scenes without being limited to a predefined set of categories. With the latest advancements in visual foundation models like SAM, it is now easier to enhance the scene understanding capabilities of AR/MR applications, leading to more intuitive and context-aware user experiences.
    \item \emph{Hand Tracking.} By efficiently capturing and rendering the fine details of hand movements and gestures in real time, 3DGS can provide more accurate and fluid hand tracking, essential for XR interactions. Also, 3DGS can be leveraged to estimate hand pose with high precision, leading to precise recognition of complex gestures and finger movements. With reduced latency and improved dynamic motion capture in hand tracking, XR gaming, remote collaboration and virtual training, can benefit.
    \item \emph{Passthrough Capabilities.} For XR devices, this function aims to blend physical and virtual worlds, improving safety and immersion given appropriate interactions between virtual content and real-world surroundings. Following a similar technical approach used for novel view synthesis, 3DGS can enhance pass-through capabilities by delivering accurate real-time reconstructions of the user's immediate environment. Simultaneously, practical challenges may arise, such as representing novel objects, maintaining high visual fidelity across various lighting conditions, \emph{etc}.
    \item \emph{Immersive Visualization.} By representing complex datasets as efficient 3D Gaussians with rich attributes, users can explore and interact with data more intuitively in immersive XR applications. 3DGS-based visualization approaches hold significant promise for fields such as scientific research, medical education, and statistical analytics, where processing and understanding complex, high-dimensional data is a fundamental problem.
\end{itemize}

\section{Conclusion}
In this paper, we explore the intersection of 3DGS and XR, synthesizing key innovations in 3DGS that hold promise for advancing XR research and development. Through a comprehensive analysis of 3DGS literature, we highlight XR research directions where 3DGS could have a transformative impact. We expect this study to inspire further 3DGS research and push the boundaries of XR. Due to the rapid progress in both fields, we may have omitted some works that could also contribute to the discussion. In future work, we aim to deliver a comprehensive survey covering the entire spectrum of 3DGS advancements and their implications for XR.

\bibliographystyle{IEEEtran}
\bibliography{egbib}

\end{document}